# Exploring Combinatorial Problem Solving with Large Language Models: A Case Study on the Travelling Salesman Problem Using GPT-3.5 Turbo

**Mahmoud Masoud[abc], Ahmed Abdelhay[d] and Mohammed Elhenawy*[1]**

[a] *Department of Information Systems & Operations Management, King Fahd University of Petroleum & Minerals, Dhahran, Saudi Arabia.*
[b] *SDAIA-KFUPM Joint Research Center for Artificial Intelligence (JRC-AI), King Fahd University of Petroleum & Minerals, Dhahran, Saudi Arabia.*

[c] *Center for Smart Mobility and Logistics, King Fahd University of Petroleum & Minerals, Dhahran, Saudi Arabia.*

[1] *Accident Research and Road Safety at the Queensland University of Technology, Australia.*

*Abstract*—Large Language Models (LLMs) are deep learning models designed to generate text based on textual input. Although researchers have been developing these models for more complex tasks such as code generation and general reasoning, few efforts have explored how LLMs can be applied to combinatorial problems. In this research, we investigate the potential of LLMs to solve the Travelling Salesman Problem (TSP). Utilizing GPT-3.5 Turbo, we conducted experiments employing various approaches, including zero-shot in-context learning, few-shot in-context learning, and chain-of-thoughts (CoT). Consequently, we fine-tuned GPT-3.5 Turbo to solve a specific problem size and tested it using a set of various instance sizes. The fine-tuned models demonstrated promising performance on problems identical in size to the training instances and generalized well to larger problems. Furthermore, to improve the performance of the fine-tuned model without incurring additional training costs, we adopted a self-ensemble approach to improve the quality of the solutions.

1. Introduction

Large language models (LLM) showed a great impact in solving many problems [1], [2]. Researchers tried a variety of methods to optimize LLMs' responses to in-context prompts. Meanwhile, one group of response optimization methods depends on engineering in-context learning prompts fed to the model by using in-context learning techniques such as (zero-shot, few-shot, CoT, etc) [4], [5], [6], [7], [8], [9], [10], other researchers investigate the potential of combining ensemble learning techniques with other in-context learning techniques to increase the optimality of the final response [8], [11], [12]. In several research papers, combining in-context learning techniques with LLM fine-tuning showed an increase in response accuracy [13], [14], [15].

Many researchers worked to solve combinatorial problems using LLM. Paper [3], presents the first study on LLMs as evolutionary combinatorial optimizers. It tests the effectiveness of using an approach referred to as LLM-driven evolutionary algorithms (LMEA) to solve TSP. LMEA instructs the LLM to select parent solutions from the existing population and perform crossover and mutation to generate offspring solutions in each generation of the evolutionary search. Then, LMEA evaluates these new solutions and includes them in the population for the next generation. This approach performs competitively to traditional heuristics in finding high-quality solutions on TSP instances with up to 20 nodes. Paper [16] suggested another approach named PROmpting (OPRO), an approach to leverage LLM as an optimizer. After describing the optimization task in natural language; In each optimization step, the LLM generates new solutions from the prompt that contains previously generated solutions with their values, after that the new solutions are evaluated and added to the prompt for the next optimization step. Meanwhile, this area of research is still under investigation, our study is the first to investigate the potential of fine-tuning LLMs to solve optimization problems, specifically TSP.

---

[1] * Corresponding Author

The main contributions of this paper are investigating the potential of using LLMs based on GPT-3.5-turbo to plan optimal drone journeys as a travel salesman problem (TSP). Investigating LLMs' performance to solve this problem will positively impact our research, as we can use LLMs to plan the drone journey. In this paper, we demonstrate our work to solve TSP using LLMs. The proposed approach will include the following steps:
- We created a dataset of simulated Journeys.
- We work on engineering in-context learning prompts with state-of-the-art prompting techniques.
- We fine-tune and validate one of the openAI text generation models (GPT3.5-turbo-0125).
- We apply and validate self-ensemble techniques over a fine-tuned model.

In addition to addressing the gap in limited studies on the use of Large Language Models (LLMs) for combinatorial problems, this paper explores whether non-expert individuals can effectively utilize off-the-shelf LLMs to solve the Travelling Salesman Problem (TSP) or similar challenges relevant to managing a small business. The following research questions drive this inquiry:
1) Can a large language model solve the Travelling Salesman Problem (TSP) based on its prior knowledge in the case of zero-shot scenarios and with a few solved examples in the case of few-shot scenarios?
2) How significantly are TSP solutions improved when the model is fine-tuned using fixed-size instances?
3) Does this fine-tuned model generalise to larger sizes of instances?
4) Does self-ensemble (i.e., self-consistency) enhance the TSP solutions of the fine-tuned models without additional fine-tuning?

2. Methodology

Our methodology, designed to address the research questions outlined in the introduction, utilizes the GPT-3.5-turbo model for this study due to its robust performance and ease of access for prompting, fine-tuning, and testing via the OpenAI API. This accessibility ensures that even individuals with minimal experience can utilize the model effectively.

To address the first research question, we employed various prompting techniques, details of which are elaborated in subsequent sections. For the second and third research questions, we fine-tuned the GPT-3.5 Turbo model using TSP instances of a fixed size and then tested it on other instances of the same training size as well as on larger sizes.

Lastly, to respond to the fourth research question, we set the model's temperature equal to 0.7 and prompted it multiple times with the same instance to assess performance enhancements achieved through self-ensemble methods.

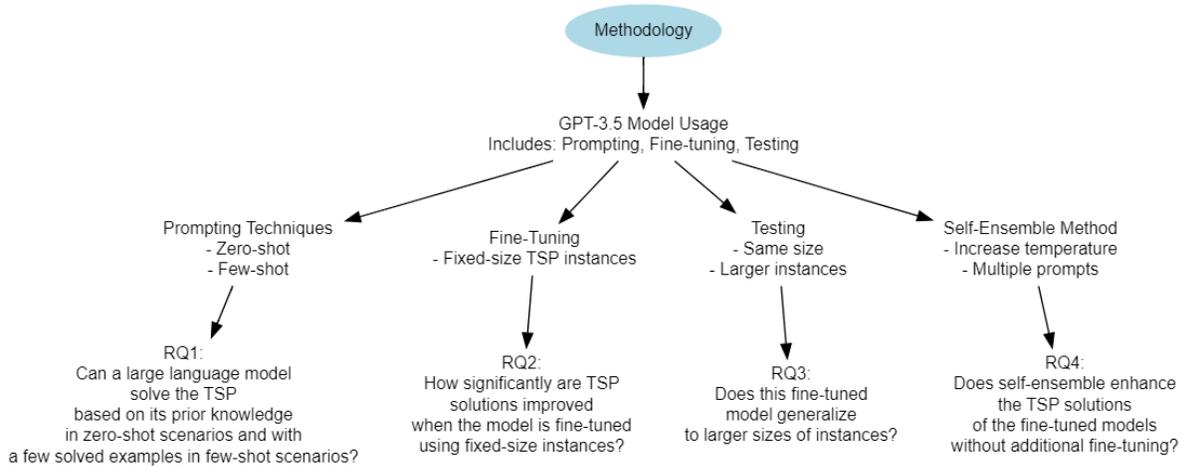

Figure 1: Methodology plan to answer research questions. Each experiment responds to one of the research questions.

*A. Dataset generation*

In this research, we used a dataset of journeys, where each journey is a number of unique 2-dimensional points. To generate this dataset, we collected journeys from the TSP-specific dataset TSPLIB and we collected randomly generated points. We used journeys from TSPLIB for in-context learning and fine-tuning; meanwhile, the randomly generated journeys were used for testing fine-tunned models. We selected the TSPLIB 95 (http://comopt.ifi.uni-heidelberg.de/software/TSPLIB95/tsp95.pdf) dataset as the base for our journey points dataset. TSPLIB is a library of sample instances for the TSP (and related problems) from various sources and types. Each file consists of a specification part and a data part. The specification part contains information on the file format and its contents. The data part contains explicit data. The journey data can be stored in various formats. We selected files with points formatted in 2-dimensional coordinates, presenting the points in the training instances. The number of points in TSPLIB journeys with the 2D point data type is variable and large. We collected 400 instances by taking the first 10 points of 400 journeys' files.

On the other hand, the testing dataset is generated randomly with a different schema. Because we want to test LLM response against journeys with different numbers of points, we created the data as follows:

- We selected this journey size {10, 11, 12, 13, 14, 15, 16, 17, 18, 19, 20, 21, 22, 27}
- For each journey size, we generate 30 different journeys that have several points equal to the journey size.

*B. In-context Learning Prompt Engineering*

Prompt engineering is an intensive task with many trials and errors. For this task, we decided to prompt GPT3.5-turbo-0125. We used many techniques to develop well-designed prompts and followed the recommended prompt-designing strategy to build our prompts. In this task, we use our generated testing data. We used four prompt formats, each applies a different combination of prompting techniques, zero-shot, zero-shot with CoT, few-shot, and few-shot with CoT. Each prompt contains context, instruction, input data, and output indicator parts. In zero-shot prompts; as shown in Table 1, we start with a context part that explains the needed knowledge to clarify the task and the formula used to calculate the distance between stations. After explaining the prompt task, we explained the expected format of the model output, and at last, we added the input data of the stations. For zero-shot with CoT in Table 2, we add a paragraph that explains who to do the task after the task explanation part. For both few-shots shown in Table 3 and few-shots with CoT shown in Table 4 prompts, we create it by adding 5 solved zero-shot or zero-shot with CoT examples of prompts and answers before the prompt we need to solve.

Table 1: Zero-shot in-context learning prompt

---

In two-dimensional space, you will visit variable number of stations. You must visit each station once and return to the starting station at the end. Each station is represented with a 2-dimensional Cartesian point ( x, y ) where x is the coordinate on the X-axis and y is the coordinate on the Y-axis. The formula calculates the Euclidean distance between stations is " ( ( X1 - X2 ) ^ 2 + ( Y1 - Y1 ) ^ 2 ) ^ 0.5 ".

Your task is to find the visiting order for the stations that minimizes the total distance you will travel to finish the journey.

The answer format should be as follows: distance matrix, Stations' order with minimum total traveling distance, and traveling distance.

station 0 ( 46, 61 ), station 1 ( 39, 56 ), station 2 ( 51, 63 ), station 3 ( 60, 91 ), station 4 ( 85, 41 ), station 5 ( 82, 83 ), station 6 ( 3, 91 ), station 7 ( 16, 67 ), station 8 ( 37, 17 ), and station 9 ( 91, 95 ).

---

Table 2: Zero-shot with CoT in-context learning prompt

---

In two-dimensional space, you will visit variable number of stations. You must visit each station once and return to the starting station at the end. Each station is represented with a 2-dimensional Cartesian point ( x, y ) where x is the coordinate on the X-axis and y is the coordinate on the Y-axis. The formula calculates the Euclidean distance between stations is " ( ( X1 - X2 ) ^ 2 + ( Y1 - Y1 ) ^ 2 ) ^ 0.5 ".

Your task is to find the visiting order for the stations that minimizes the total distance you will travel to finish the journey.

Let's work this out step-by-step to ensure we have the correct answer. First, calculate the Euclidean distance between all pairs of stations using the Euclidean distance formula and make a matrix of the calculated distances. Then, compare all the possible stations' orders to find the order that costs the minimum total traveling distance of the journey. At last, Sum the distances between the stations according to the order you find.

The answer format should be as follows: distance matrix, Stations' order with minimum total traveling distance, and traveling distance.

station 0 ( 46, 61 ), station 1 ( 39, 56 ), station 2 ( 51, 63 ), station 3 ( 60, 91 ), station 4 ( 85, 41 ), station 5 ( 82, 83 ), station 6 ( 3, 91 ), station 7 ( 16, 67 ), station 8 ( 37, 17 ), and station 9 ( 91, 95 ).

---

Table 3: Few-shot in-context learning prompt

---

In two-dimensional space, you will visit variable number of stations. You must visit each station once and return to the starting station at the end. Each station is represented with a 2-dimensional Cartesian point ( x, y ) where x is the coordinate on the X-axis and y is the coordinate on the Y-axis. The formula calculates the Euclidean distance between stations is " ( ( X1 - X2 ) ^ 2 + ( Y1 - Y1 ) ^ 2 ) ^ 0.5 ".

Your task is to find the visiting order for the stations that minimizes the total distance you will travel to finish the journey.Let's work this out step-by-step to ensure we have the correct answer. First, calculate the Euclidean distance between all pairs of stations using the Euclidean distance formula and make a matrix of the calculated distances. Then, compare all the possible stations' orders to find the order that costs the minimum total traveling distance of the journey. At last, Sum the distances between the stations according to the order you find.

The answer format should strictly follows a json format with no description as follows: distance_matrix, Stations' order with minimum total traveling distance as minimum_distance_order, and traveling_cost.

station 0 ( 4, 5 ), station 1 ( 5, 6 ), station 2 ( 1, 1 ), station 3 ( 5, 6 ), station 4 ( 1, 9 ), station 5 ( 1, 1 ), station 6 ( 7, 5 ), station 7 ( 7, 8 ), station 8 ( 4, 3 ), and station 9 ( 1, 2 ).

{"distance_matrix":"[[0, 1.41, 5, 1.41, 5, 5, 3, 4.24, 2, 4.24], [1.41, 0, 6.4, 0, 5, 6.4, 2.24, 2.83, 3.16, 5.66], [5, 6.4, 0, 6.4, 8, 0, 7.21, 9.22, 3.61, 1], [1.41, 0, 6.4, 0, 5, 6.4, 2.24, 2.83, 3.16, 5.66], [5, 5, 8, 5, 0, 8, 7.21, 6.08, 6.71, 7], [5, 6.4, 0, 6.4, 8, 0, 7.21, 9.22, 3.61, 1], [3, 2.24,

7.21, 2.24, 7.21, 7.21, 0, 3, 3.61, 6.71], [4.24, 2.83, 9.22, 2.83, 6.08, 9.22, 3, 0, 5.83, 8.49], [2, 3.16, 3.61, 3.16, 6.71, 3.61, 3.61, 5.83, 0, 3.16], [4.24, 5.66, 1, 5.66, 7, 1, 6.71, 8.49, 3.16, 0]]"}

station 0 ( 3, 1 ), station 1 ( 4, 1 ), station 2 ( 6, 6 ), station 3 ( 8, 3 ), station 4 ( 0, 9 ), station 5 ( 8, 9 ), station 6 ( 3, 3 ), station 7 ( 3, 1 ), station 8 ( 2, 9 ), and station 9 ( 5, 0 ).

{"distance_matrix":"[[0, 1, 5.83, 5.39, 8.54, 9.43, 2, 0, 8.06, 2.24], [1, 0, 5.39, 4.47, 8.94, 8.94, 2.24, 1, 8.25, 1.41], [5.83, 5.39, 0, 3.61, 6.71, 3.61, 4.24, 5.83, 5, 6.08], [5.39, 4.47, 3.61, 0, 10, 6, 5, 5.39, 8.49, 4.24], [8.54, 8.94, 6.71, 10, 0, 8, 6.71, 8.54, 2, 10.3], [9.43, 8.94, 3.61, 6, 8, 0, 7.81, 9.43, 6, 9.49], [2, 2.24, 4.24, 5, 6.71, 7.81, 0, 2, 6.08, 3.61], [0, 1, 5.83, 5.39, 8.54, 9.43, 2, 0, 8.06, 2.24], [8.06, 8.25, 5, 8.49, 2, 6, 6.08, 8.06, 0, 9.49], [2.24, 1.41, 6.08, 4.24, 10.3, 9.49, 3.61, 2.24, 9.49, 0]]"}

station 0 ( 5, 1 ), station 1 ( 0, 0 ), station 2 ( 4, 8 ), station 3 ( 3, 4 ), station 4 ( 6, 0 ), station 5 ( 1, 2 ), station 6 ( 9, 4 ), station 7 ( 8, 8 ), station 8 ( 5, 1 ), and station 9 ( 6, 5 ).

{"distance_matrix":"[[0, 5.1, 7.07, 3.61, 1.41, 4.12, 5, 7.62, 0, 4.12], [5.1, 0, 8.94, 5, 6, 2.24, 9.85, 11.31, 5.1, 7.81], [7.07, 8.94, 0, 4.12, 8.25, 6.71, 6.4, 4, 7.07, 3.61], [3.61, 5, 4.12, 0, 5, 2.83, 6, 6.4, 3.61, 3.16], [1.41, 6, 8.25, 5, 0, 5.39, 5, 8.25, 1.41, 5], [4.12, 2.24, 6.71, 2.83, 5.39, 0, 8.25, 9.22, 4.12, 5.83], [5, 9.85, 6.4, 6, 5, 8.25, 0, 4.12, 5, 3.16], [7.62, 11.31, 4, 6.4, 8.25, 9.22, 4.12, 0, 7.62, 3.61], [0, 5.1, 7.07, 3.61, 1.41, 4.12, 5, 7.62, 0, 4.12], [4.12, 7.81, 3.61, 3.16, 5, 5.83, 3.16, 3.61, 4.12, 0]]"}

station 0 ( 7, 3 ), station 1 ( 7, 8 ), station 2 ( 5, 3 ), station 3 ( 8, 6 ), station 4 ( 7, 9 ), station 5 ( 4, 4 ), station 6 ( 9, 1 ), station 7 ( 1, 7 ), station 8 ( 2, 7 ), and station 9 ( 6, 4 ).

{"distance_matrix":"[[0, 5, 2, 3.16, 6, 3.16, 2.83, 7.21, 6.4, 1.41], [5, 0, 5.39, 2.24, 1, 5, 7.28, 6.08, 5.1, 4.12], [2, 5.39, 0, 4.24, 6.32, 1.41, 4.47, 5.66, 5, 1.41], [3.16, 2.24, 4.24, 0, 3.16, 4.47, 5.1, 7.07, 6.08, 2.83], [6, 1, 6.32, 3.16, 0, 5.83, 8.25, 6.32, 5.39, 5.1], [3.16, 5, 1.41, 4.47, 5.83, 0, 5.83, 4.24, 3.61, 2], [2.83, 7.28, 4.47, 5.1, 8.25, 5.83, 0, 10, 9.22, 4.24], [7.21, 6.08, 5.66, 7.07, 6.32, 4.24, 10, 0, 1, 5.83], [6.4, 5.1, 5, 6.08, 5.39, 3.61, 9.22, 1, 0, 5], [1.41, 4.12, 1.41, 2.83, 5.1, 2, 4.24, 5.83, 5, 0]]"}

station 0 ( 7, 6 ), station 1 ( 7, 5 ), station 2 ( 5, 8 ), station 3 ( 8, 5 ), station 4 ( 7, 3 ), station 5 ( 4, 1 ), station 6 ( 9, 4 ), station 7 ( 1, 6 ), station 8 ( 2, 0 ), and station 9 ( 0, 7 ).

{"distance_matrix":"[[0, 1, 2.83, 1.41, 3, 5.83, 2.83, 6, 7.81, 7.07], [1, 0, 3.61, 1, 2, 5, 2.24, 6.08, 7.07, 7.28], [2.83, 3.61, 0, 4.24, 5.39, 7.07, 5.66, 4.47, 8.54, 5.1], [1.41, 1, 4.24, 0, 2.24, 5.66, 1.41, 7.07, 7.81, 8.25], [3, 2, 5.39, 2.24, 0, 3.61, 2.24, 6.71, 5.83, 8.06], [5.83, 5, 7.07, 5.66, 3.61, 0, 5.83, 5.83, 2.24, 7.21], [2.83, 2.24, 5.66, 1.41, 2.24, 5.83, 0, 8.25, 8.06, 9.49], [6, 6.08, 4.47, 7.07, 6.71, 5.83, 8.25, 0, 6.08, 1.41], [7.81, 7.07, 8.54, 7.81, 5.83, 2.24, 8.06, 6.08, 0, 7.28], [7.07, 7.28, 5.1, 8.25, 8.06, 7.21, 9.49, 1.41, 7.28, 0]]"}

station 0 ( 1, 8 ), station 1 ( 9, 3 ), station 2 ( 6, 8 ), station 3 ( 8, 4 ), station 4 ( 9, 5 ), station 5 ( 5, 1 ), station 6 ( 6, 2 ), station 7 ( 2, 1 ), station 8 ( 3, 6 ), and station 9 ( 6, 4 ).

Table 4: Few-shot with CoT in-context learing prompt

In two-dimensional space, you will visit variable number of stations. You must visit each station once and return to the starting station at the end. Each station is represented with a 2-dimensional Cartesian point ( x, y ) where x is the coordinate on the X-axis and y is the coordinate on the Y-axis. The formula calculates the Euclidean distance between stations is " ( ( X1 - X2 ) ^ 2 + ( Y1 - Y1 ) ^ 2 ) ^ 0.5 ".

Your task is to find the visiting order for the stations that minimizes the total distance you will travel to finish the journey.Let's work this out step-by-step to ensure we have the correct answer. First, calculate the Euclidean distance between all pairs of stations using the Euclidean distance formula and make a matrix of the calculated distances. Then, compare all the possible stations' orders to find the order that costs the minimum total traveling distance of the journey. At last, Sum the distances between the stations according to the order you find.

The answer format should strictly follows a json format with no description as follows: distance_matrix, Stations' order with minimum total traveling distance as minimum_distance_order, and traveling_cost.

station 0 ( 4, 5 ), station 1 ( 5, 6 ), station 2 ( 1, 1 ), station 3 ( 5, 6 ), station 4 ( 1, 9 ), station 5 ( 1, 1 ), station 6 ( 7, 5 ), station 7 ( 7, 8 ), station 8 ( 4, 3 ), and station 9 ( 1, 2 ).

{"distance_matrix":"[[0, 1.41, 5, 1.41, 5, 5, 3, 4.24, 2, 4.24], [1.41, 0, 6.4, 0, 5, 6.4, 2.24, 2.83, 3.16, 5.66], [5, 6.4, 0, 6.4, 8, 0, 7.21, 9.22, 3.61, 1], [1.41, 0, 6.4, 0, 5, 6.4, 2.24, 2.83, 3.16, 5.66], [5, 5, 8, 5, 0, 8, 7.21, 6.08, 6.71, 7], [5, 6.4, 0, 6.4, 8, 0, 7.21, 9.22, 3.61, 1], [3, 2.24, 7.21, 2.24, 7.21, 7.21, 0, 3, 3.61, 6.71], [4.24, 2.83, 9.22, 2.83, 6.08, 9.22, 3, 0, 5.83, 8.49], [2, 3.16, 3.61, 3.16, 6.71, 3.61, 3.61, 5.83, 0, 3.16], [4.24, 5.66, 1, 5.66, 7, 1, 6.71, 8.49, 3.16, 0]]"}

station 0 ( 3, 1 ), station 1 ( 4, 1 ), station 2 ( 6, 6 ), station 3 ( 8, 3 ), station 4 ( 0, 9 ), station 5 ( 8, 9 ), station 6 ( 3, 3 ), station 7 ( 3, 1 ), station 8 ( 2, 9 ), and station 9 ( 5, 0 ).

{"distance_matrix":"[[0, 1, 5.83, 5.39, 8.54, 9.43, 2, 0, 8.06, 2.24], [1, 0, 5.39, 4.47, 8.94, 8.94, 2.24, 1, 8.25, 1.41], [5.83, 5.39, 0, 3.61, 6.71, 3.61, 4.24, 5.83, 5, 6.08], [5.39, 4.47, 3.61, 0, 10, 6, 5, 5.39, 8.49, 4.24], [8.54, 8.94, 6.71, 10, 0, 8, 6.71, 8.54, 2, 10.3], [9.43, 8.94, 3.61, 6, 8, 0, 7.81, 9.43, 6, 9.49], [2, 2.24, 4.24, 5, 6.71, 7.81, 0, 2, 6.08, 3.61], [0, 1, 5.83, 5.39, 8.54, 9.43, 2, 0, 8.06, 2.24], [8.06, 8.25, 5, 8.49, 2, 6, 6.08, 8.06, 0, 9.49], [2.24, 1.41, 6.08, 4.24, 10.3, 9.49, 3.61, 2.24, 9.49, 0]]"}

station 0 ( 5, 1 ), station 1 ( 0, 0 ), station 2 ( 4, 8 ), station 3 ( 3, 4 ), station 4 ( 6, 0 ), station 5 ( 1, 2 ), station 6 ( 9, 4 ), station 7 ( 8, 8 ), station 8 ( 5, 1 ), and station 9 ( 6, 5 ).

{"distance_matrix":"[[0, 5.1, 7.07, 3.61, 1.41, 4.12, 5, 7.62, 0, 4.12], [5.1, 0, 8.94, 5, 6, 2.24, 9.85, 11.31, 5.1, 7.81], [7.07, 8.94, 0, 4.12, 8.25, 6.71, 6.4, 4, 7.07, 3.61], [3.61, 5, 4.12, 0, 5, 2.83, 6, 6.4, 3.61, 3.16], [1.41, 6, 8.25, 5, 0, 5.39, 5, 8.25, 1.41, 5], [4.12, 2.24, 6.71, 2.83, 5.39, 0, 8.25, 9.22, 4.12, 5.83], [5, 9.85, 6.4, 6, 5, 8.25, 0, 4.12, 5, 3.16], [7.62, 11.31, 4, 6.4, 8.25, 9.22, 4.12, 0, 7.62, 3.61], [0, 5.1, 7.07, 3.61, 1.41, 4.12, 5, 7.62, 0, 4.12], [4.12, 7.81, 3.61, 3.16, 5, 5.83, 3.16, 3.61, 4.12, 0]]"}

station 0 ( 7, 3 ), station 1 ( 7, 8 ), station 2 ( 5, 3 ), station 3 ( 8, 6 ), station 4 ( 7, 9 ), station 5 ( 4, 4 ), station 6 ( 9, 1 ), station 7 ( 1, 7 ), station 8 ( 2, 7 ), and station 9 ( 6, 4 ).

{"distance_matrix":"[[0, 5, 2, 3.16, 6, 3.16, 2.83, 7.21, 6.4, 1.41], [5, 0, 5.39, 2.24, 1, 5, 7.28, 6.08, 5.1, 4.12], [2, 5.39, 0, 4.24, 6.32, 1.41, 4.47, 5.66, 5, 1.41], [3.16, 2.24, 4.24, 0, 3.16, 4.47, 5.1, 7.07, 6.08, 2.83], [6, 1, 6.32, 3.16, 0, 5.83, 8.25, 6.32, 5.39, 5.1], [3.16, 5, 1.41, 4.47, 5.83, 0, 5.83, 4.24, 3.61, 2], [2.83, 7.28, 4.47, 5.1, 8.25, 5.83, 0, 10, 9.22, 4.24], [7.21, 6.08, 5.66, 7.07, 6.32, 4.24, 10, 0, 1, 5.83], [6.4, 5.1, 5, 6.08, 5.39, 3.61, 9.22, 1, 0, 5], [1.41, 4.12, 1.41, 2.83, 5.1, 2, 4.24, 5.83, 5, 0]]"}

station 0 ( 7, 6 ), station 1 ( 7, 5 ), station 2 ( 5, 8 ), station 3 ( 8, 5 ), station 4 ( 7, 3 ), station 5 ( 4, 1 ), station 6 ( 9, 4 ), station 7 ( 1, 6 ), station 8 ( 2, 0 ), and station 9 ( 0, 7 ).

{"distance_matrix":"[[0, 1, 2.83, 1.41, 3, 5.83, 2.83, 6, 7.81, 7.07], [1, 0, 3.61, 1, 2, 5, 2.24, 6.08, 7.07, 7.28], [2.83, 3.61, 0, 4.24, 5.39, 7.07, 5.66, 4.47, 8.54, 5.1], [1.41, 1, 4.24, 0, 2.24, 5.66, 1.41, 7.07, 7.81, 8.25], [3, 2, 5.39, 2.24, 0, 3.61, 2.24, 6.71, 5.83, 8.06], [5.83, 5, 7.07, 5.66, 3.61, 0, 5.83, 5.83, 2.24, 7.21], [2.83, 2.24, 5.66, 1.41, 2.24, 5.83, 0, 8.25, 8.06, 9.49], [6, 6.08, 4.47, 7.07, 6.71, 5.83, 8.25, 0, 6.08, 1.41], [7.81, 7.07, 8.54, 7.81, 5.83, 2.24, 8.06, 6.08, 0, 7.28], [7.07, 7.28, 5.1, 8.25, 8.06, 7.21, 9.49, 1.41, 7.28, 0]]"}

station 0 ( 1, 8 ), station 1 ( 9, 3 ), station 2 ( 6, 8 ), station 3 ( 8, 4 ), station 4 ( 9, 5 ), station 5 ( 5, 1 ), station 6 ( 6, 2 ), station 7 ( 2, 1 ), station 8 ( 3, 6 ), and station 9 ( 6, 4 ).

*C. Fine-Tuning*

We used our training dataset of 400 examples to fine-tune openAI's GPT3.5-turbo-0125 model. We used OpenAI's fine-tuning GUI on their official website. OpenAI requires using JSONL format to feed the training dataset and start fine-tuning tasks. Table 5 shows OpenAI's template for training examples. Each example consists of 3 messages, each with a role property and a content property. Possible roles are system role, user role, and assistant role. The message with system role, contains task's context, instruction, output indicator, and CoT section if existing. The message with the user role contains the input data of journey stations. The last message with an assistant role, contains an example of the response we expect the model to generate. Also, OpenAI requires to set all training example in one JSONL file. Also, each training example should be in one line.

Table 5: OpenAI Fine-tuning prompt template

{"messages":[{"role":"system","content":"In two-dimensional space, you will visit 10 stations. You must visit each station once and return to the starting station at the end. Each station is represented with a 2-dimensional Cartesian point ( x, y ) where x is the coordinate on the X-axis and y is the coordinate on the Y-axis. The formula calculates the Euclidean distance between stations is \" ( ( X1 - X2 ) ^ 2 + ( Y1 - Y1 ) ^ 2 ) ^ 0.5 \".\n\nYour task is to find the visiting order for the stations that minimizes the total distance you will travel to finish the journey.\n\nLet's work this out step-by-step to ensure we have the correct answer. First, calculate the Euclidean distance between all pairs of stations using the Euclidean distance formula and make a matrix of the calculated distances. Then, compare all the possible stations' orders to find the order that costs the minimum total travelling distance of the journey. At last, Sum the distances between the stations according to the order you find.\n\nThe answer format should be as follows: distance matrix, Stations' order with minimum total traveling distance, and traveling distance.\n"},{"role":"user","content":"station 0 ( 8775, 9800 ), station 1 ( 8575, 9800 ), station 2 ( 8375, 10500 ), station 3 ( 8775, 10450 ), station 4 ( 8375, 10700 ), station 5 ( 8175, 10700 ), station 6 ( 8775, 10650 ), station 7 ( 8575, 10650 ), station 8 ( 8375, 11300 ), and station 9 ( 8775, 11300 )."},{"role":"assistant","content":{"distance_matrix":[[0,200,806.23,650,984.89,1081.67,850,873.21,1552.42,1500],[200,0,728.01,680.07,921.95,984.89,873.21,850,1513.27,1513.27],[806.23,728.01,0,403.11,200,282.84,427.2,250,800,894.43],[650,680.07,403.11,0,471.7,650,200,282.84,939.41,850],[984.89,921.95,200,471.7,0,200,403.11,206.16,600,721.11],[1081.67,984.89,282.84,650,200,0,602.08,403.11,632.46,848.53],[850,873.21,427.2,200,403.11,602.08,0,200,763.22,650],[873.21,850,250,282.84,206.16,403.11,200,0,680.07,680.07],[1552.42,1513.27,800,939.41,600,632.46,763.22,680.07,0,400],[1500,1513.27,894.43,850,721.11,848.53,650,680.07,400,0]],"minimum_distance_order":[0,1,2,4,5,8,9,7,6,3,0],"traveling_cost":4090.54}}]}

*D. Check Model Response Hallucination*

In the context of TSP, "hallucination" refers to instances where the model outputs journeys that violate the Hamiltonian path condition, which requires visiting each city exactly once and returning to the starting point. A typical hallucination observed in our study, especially at temperatures higher than 0.7, involves the model revisiting certain nodes, such as nodes3 and 10, while completely omitting others like nodes1 and 2, as demonstrated in the example provided in Table 6. This type of error has been the most common hallucination encountered, guiding us to adhere to the recommended temperature for self-ensemble as suggested by [8] in their study on enhancing reasoning in language models through self-consistency (i.e. self-ensemble). To check hallucination, we do the following:

A. Extract the order of points returned in the response using this regular expression ""(?:minimum_distance_|optimal_)?(?:order|path)"\s*:\s*\[(\s*\d+(?:,\s*\d+)*\s*)\]"

B. If order exist, we check for miss-visiting any point, multiple visits to points, or not returning to starting point at the end of the order.

C. If there are no miss-visits, multiple visits, and the order return to starting point, then the order has no hallucination.

Table 6: Model hallucinates by doing multiple visits for {3, 10} and mis visit {1, 2}

| Prompts |
|---|
| 'In two-dimensional space, you will visit variable number of stations. You must visit each station once and return to the starting station at the end. Each station is represented with a 2-dimensional Cartesian point ( x, y ) where x is the coordinate on the X-axis and y is the coordinate on the Y-axis. The formula calculates the Euclidean distance between stations is " ( ( X1 - X2 ) ^ 2 + ( Y1 - Y1 ) ^ 2 ) ^ 0.5 ". |
| 'Your task is to find the visiting order for the stations that minimizes the total distance you will travel to finish the journey. |
| "Let's work this out step-by-step to ensure we have the correct answer. First, calculate the Euclidean distance between all pairs of stations using the Euclidean distance formula and make a matrix of the calculated distances. Then, compare all the possible stations' orders to find the order that costs the minimum total travelling distance of the journey. At last, Sum the distances between the stations according to the order you find. |
| "The answer format should be as follows: distance matrix, Stations' order with minimum total traveling distance, and traveling distance. |
| station 0 ( 9470379, 8803158 ), station 1 ( 10428106, 15842209 ), station 2 ( 11637181, 18854782 ), station 3 ( 9118, 7514 ), station 4 ( 604825, 401373 ), station 5 ( 5071, 6491 ), station 6 ( 91363, 117545 ), station 7 ( 71806, 54776 ), station 8 ( 1699973, 2422256 ), station 9 ( 24525, 36301 ), and station 10 ( 39580, 37868 ). |
| **Response** |
| "{\"distance_matrix\":[[\"0.00\",\"7151748.21\",\"10915688.37\",\"9470450.34\",\"8890192.99\",\"9469700.30\",\"9447653.50\",\"9408331.66\",\"7452620.91\",\"9460033.42\",\"9460441.02\"],[\"7151748.21\",\"0.00\",\"3506732.15\",\"15870664.62\",\"15629331.05\",\"15872891.75\",\"15847310.40\",\"15831358.10\",\"14274609.53\",\"15826130.55\",\"15824816.99\"],[\"10915688.37\",\"3506732.15\",\"0.00\",\"18747330.78\",\"18506888.57\",\"18750090.57\",\"18724905.73\",\"18708650.77\",\"17151489.33\",\"18701469.56\",\"18700171.05\"],[\"9470450.34\",\"15870664.62\",\"18747330.78\",\"0.00\",\"443694.41\",\"2617.75\",\"158682.79\",\"101044.38\",\"2232596.26\",\"5012.90\",\"2276.20\"],[\"8890192.99\",\"15629331.05\",\"18506888.57\",\"443694.41\",\"0.00\",\"505437.54\",\"358928.12\",\"331309.36\",\"1583060.35\",\"370196.69\",\"354516.12\"],[\"9469700.30\",\"15872891.75\",\"18750090.57\",\"2617.75\",\"505437.54\",\"0.00\",\"117878.86\",\"72090.05\",\"1694988.41\",\"29358.57\",\"21607.49\"],[\"9447653.50\",\"15847310.40\",\"18724905.73\",\"158682.79\",\"358928.12\",\"117878.86\",\"0.00\",\"63070.71\",\"1651482.13\",\"82108.42\",\"75204.83\"],[\"9408331.66\",\"15831358.10\",\"18708650.77\",\"101044.38\",\"331309.36\",\"72090.05\",\"63070.71\",\"0.00\",\"1558830.50\",\"48535.88\",\"47068.51\"],[\"7452620.91\",\"14274609.53\",\"17151489.33\",\"2232596.26\",\"1583060.35\",\"1694988.41\",\"1651482.13\",\"1558830.50\",\"0.00\",\"1672467.38\",\"1671088.32\"],[\"9460033.42\",\"15826130.55\",\"18701469.56\",\"5012.90\",\"370196.69\",\"29358.57\",\"82108.42\",\"48535.88\",\"1672467.38\",\"0.00\",\"1784.89\"],[\"9460441.02\",\"15824816.99\",\"18700171.05\",\"2276.20\",\"354516.12\",\"21607.49\",\"75204.83\",\"47068.51\",\"1671088.32\",\"1784.89\",\"0.00\"]],\"minimum_distance_order\":[0,==3==,5,==10==,9,4,6,7,8,==10==,==3==,0],\"traveling_cost\":37401513.78580033}" |

## E. Check Response Accuracy

In this research, we evaluated the TSP solutions using two distinct measures. The first measure, which we refer to as the randomness score, ranges from zero to one. As shown in Figure 2 to calculate this score, we first randomly reorder the cities in a TSP instance—effectively generating a random solution—and then evaluate the cost of this random itinerary. By repeating this randomization process and cost evaluation, we create a distribution of costs for randomly generated solutions. We then compare the cost of the model's solution to this distribution. Specifically, in algorithm 1 we count the number of random solutions that have a cost less than or equal to that of the model's solution and divide this number by the total number of random solutions generated. This score allows us to test the null hypothesis that the model returns a random ordering of cities. The randomness score effectively acts as the p-value for the solution being generated randomly. If the score is less than 0.05, we reject the null hypothesis and conclude that the solution is not randomly generated. Conversely, a score higher than this threshold indicates that the model's solution might be random. By repeating this procedure across multiple TSP instances, we can estimate the likelihood that the model returns random solutions.

The second measure, known as the gap, algorithm 2 calculates the difference between the cost of the solution provided by the model and the optimal cost, normalised by the optimal cost. This widely recognized metric helps assess the efficiency of the proposed solution relative to the best possible outcome.

Algorithm 1: Randomness check

Input: Model order cost $C_M$, random orders cost array $Arr\_C_R$, number of random orders N;

Output: randomness score P;

1. $C_{R-valid} \leftarrow \{C_R \in Arr\_C_R \mid C_R \leq C_M\}$;
2. $P \leftarrow \frac{C_{R-valid}}{N}$;

return P;

Algorithm 2: Gap calculation

Input: Model order cost $C_M$, optimal order cost $C_o$;

Output: gap value G;

1. $G \leftarrow \frac{C_M - C_O}{C_O}$;

return G;

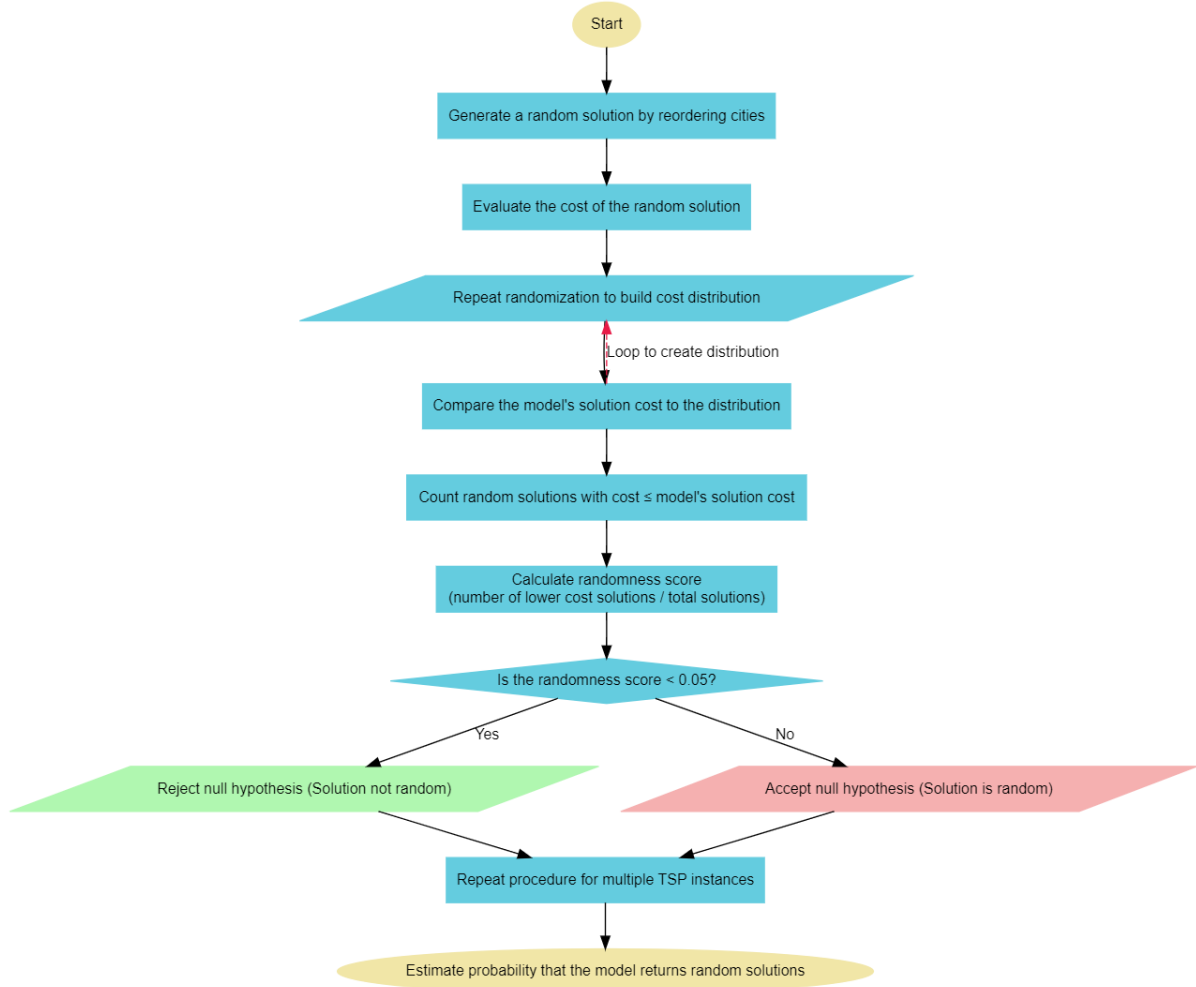

Figure 2: Model hallucinates by doing multiple visits for {3, 10} and mis visit {1, 2}

## F. Self-Ensemble

To implement self-ensemble in this study, we prompted the model eleven times with a temperature of 0.7 to obtain multiple responses for each journey in the test dataset. We then assessed the effect of self-ensemble using varying ensemble sizes: {1,3,5,7,9,11}. For instance, to determine a solution for an ensemble of size B, we would select the first B responses from the eleven obtained. Next, we filtered out any responses that were deemed to be hallucination, we calculated the cost of each solution and selected the one with the lowest cost as the definitive solution for that particular instance as shown in algorithm 3.

Algorithm 3: Self-Ensemble

---

Input: The set of all responses for a particular journey in the test dataset Response_Arr, ensemble size B;

Output: The best response $r^*$ and its cost;

Definitions:

$B \in \{1,3,5,7,9,11\}$;

C(r) is the cost function that assigns a cost to each response r;

isHallucination(r) is a predicate function that evaluates to true if the response r is considered a hallucination;

1. $Response\_Arr_B \leftarrow \{r_i \mid r_i \in Response\_Arr, i = 1, \dots, B\}$;
2. $Response\_Arr_{valid} \leftarrow \{r \in Response\_Arr_B \mid \neg isHallucination(r)\}$
3. $r^* \leftarrow argmin_{r \in Response\_Arr_{valid}} C(r)$;

Return $Response\_Arr_{best} \leftarrow \{r \in Response\_Arr_B \mid r = r^*\}, C(r^*)$;

---

## 3. Experimental results

In this section, we will discuss the outcomes of our experiments. Initially, we analyse the percentage of correct trips returned by both the in-context learning techniques and the fine-tuned model, without considering the quality of these trips. A correct trip is characterised by one that begins and ends at the same location and visits each node exactly once. Subsequently, we step into the quality of these correct trips for both the in-context learning scenarios and the fine-tuned model. Given the fine-tuned model's good performance, we further examine how self-ensemble techniques enhance the proportion of correct trips returned for each instance size and boost the overall quality of the journeys.

### A. In context learning Hallucination Responses

In Figure 3 the bar chart illustrates the results from analysing 30 TSP instances across sizes ranging from 10 to 22, highlighting the correlation between instance size and the percentage of correct, non-hallucinated, responses from a language model employing various in-context learning strategies, such as zero-shot and few-shot learning, both with and without the implementation of CoT techniques. The success rates display remarkable differences between the approaches. Few-shot and few-shot with CoT techniques show promising accuracy for smaller instances, particularly those close to the size of 10. Yet, as the instance sizes increase, there is an observed increase in errors, which could be attributed to the limitations on the model's total token count. Consequently, the few-shot approach appears to uphold precision more consistently, even as instance sizes grow, underlining the benefit of providing the model with a set of examples—assuming the constraints of maximum token capacity are met. These findings show the significance of prompt engineering, which proves important not only for generating accurate responses but also for adhering to the model's token limitations.

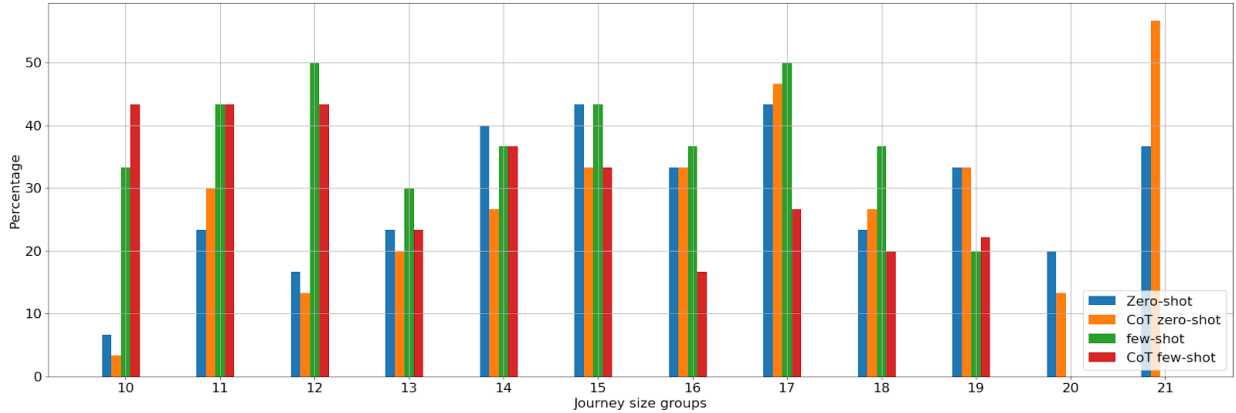

Figure 3: Good order percentage for multiple groups of similar journey size and different prompting technique

B. *Fine-tuned GPT-3.5 Hallucination Responses*

In Figure 4 the bar graph depicts the GPT-3.5 model's performance, fine-tuned on TSP instances of size 10, solving 30 instances at varying sizes. Each instance has 11 self-ensemble responses. By evaluating only, the first response (i.e., ensemble size of one), fine-tuned model exhibited a high success rate, especially for instance sizes near the fine-tuning benchmark of 10, indicating robust performance in scenarios similar to its training. As instance sizes increase, the percentage of correct trips declined, signaling the model's difficulty in scaling its solution accuracy for larger instances. Yet, it's worth noting that the fine-tuned model outperformed in-context learning methods across all instance sizes. The data also show self-ensemble's positive impact, enhancing the accuracy of correct trips. The model reliably produced at least one accurate, non-hallucinated trip for smaller instances, with the self-ensemble's advantage becoming more distinct as instance size grew; for example, for size 27, a bigger ensemble number improved the rate of correct, non-hallucinated solutions. An increasing trend of hallucinated responses for larger instance sizes, notably at 21, 22, and 27, highlights the model's escalating challenge in preserving high percentage of correct trips for larger problem instances.

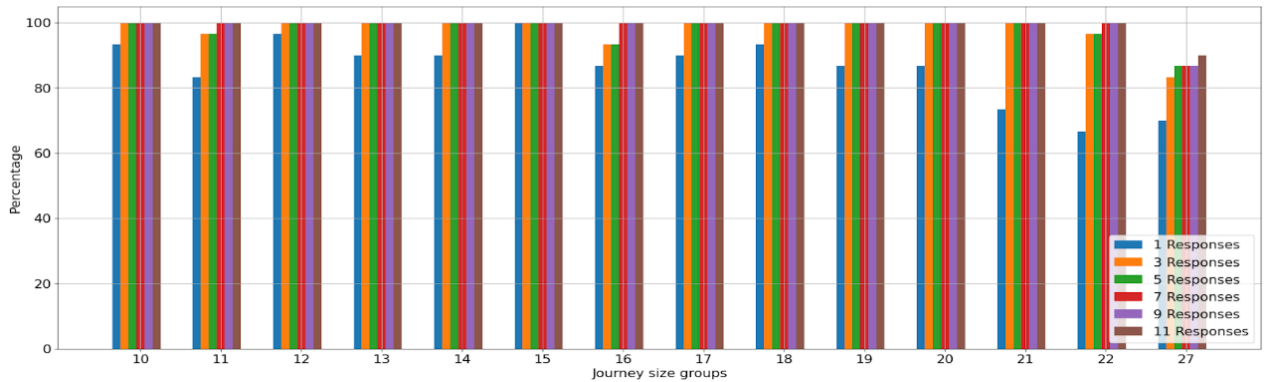

Figure 4: Compare model responses at different journey sizes with different self-ensemble points

C. *In-context trip quality analysis*

Figure 5(A) depicts the median gap values across various in-context learning techniques, offering a clear visualisation of the model's performance as the instance size increases. Regardless of the approach—whether zero-shot, zero-shot with the addition of chain of thought (COT), few-shot, or few-shot with COT—all exhibit a consistent upward trend in the median gap. This trend suggests that the challenges associated with solving the TSP grow as the problem size grows. Interestingly, incorporating COT does not appear to significantly alter this trajectory, indicating that the complexity of larger instances presents a uniform challenge across different techniques. We observe a different behaviour when Figure 5(B) presents the Interquartile Range (IQR) for the gap across the same techniques and instance sizes. Unlike the median gap, the IQR does not follow a discernible pattern of increase with the instance size. The fluctuation in the IQR suggests variability in the model's performance consistency. It is evident that while the median performance deteriorates as the instances grow, the variation in performance measured by the IQR does not align with a straightforward trend, indicating that certain instance sizes may present unique challenges or advantages that are not solely dependent on the size of the problem.

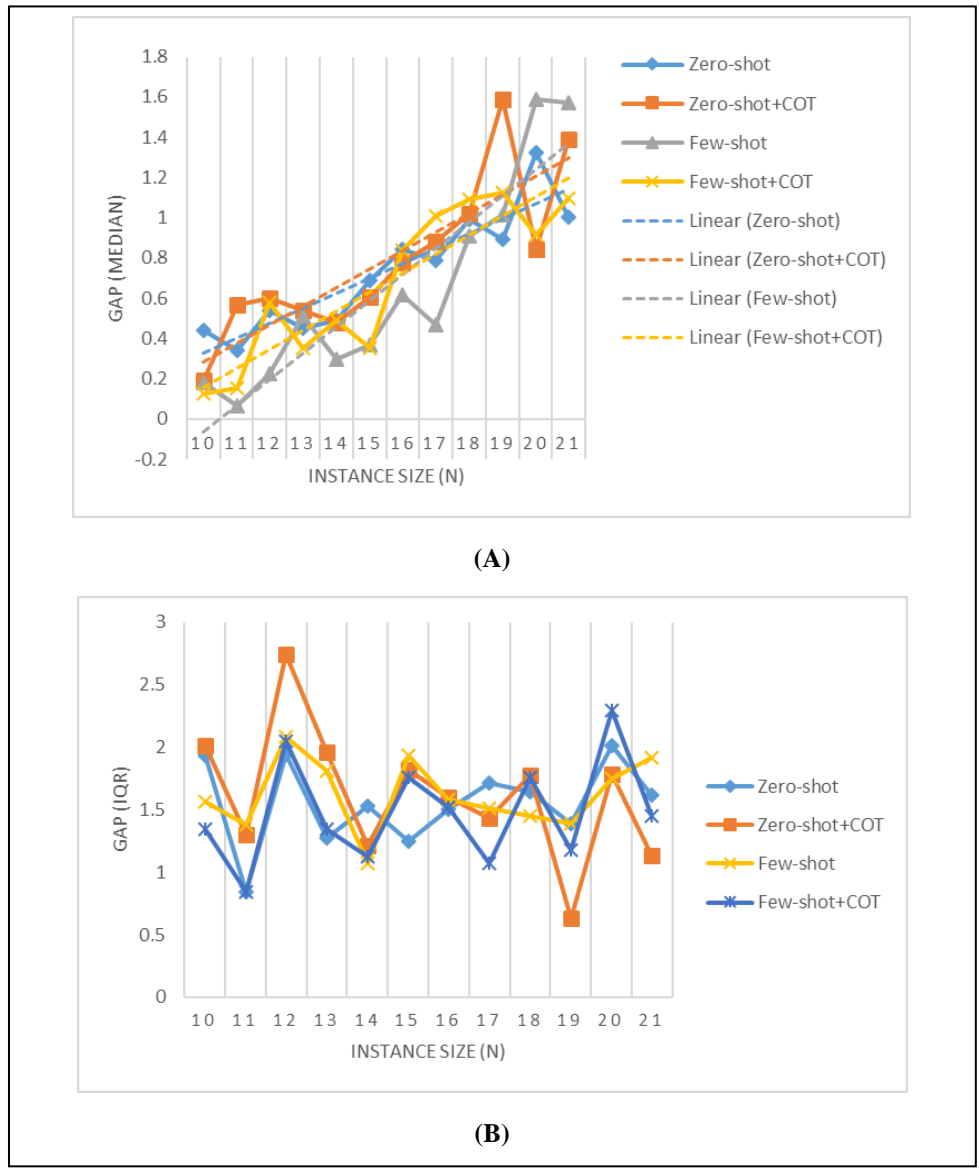

Figure 5: Median and interquartile for the gap of each group of journeys with similar size and different in-context prompting technique for gpt3.5-turbo-0125

Table 7 provides randomness scores for solutions to the Travelling Salesman Problem (TSP) spanning various instance sizes. These scores compare zero-shot and few-shot in-context learning techniques, with and without the integration of Chain of Thought (CoT). To evaluate the randomness of the solutions, the median scores and their Interquartile Ranges (IQR) are analysed. The median scores reveal that zero-shot approaches frequently result in higher randomness scores, often surpassing the 0.05 threshold. This suggests the solutions might be random, and we cannot reject the Null hypothesis, which posits that the model generates random solutions. Contrary to expectations, the addition of CoT to zero-shot learning did not significantly decrease the randomness scores. In contrast, few-shot methods generally produce lower median randomness scores compared to zero-shot, indicating a tendency toward less random solutions. Similarly, the incorporation of CoT with few-shot learning did not lead to improved randomness scores. Notably, as instance sizes grow, scores for the larger instances of 22 and 27 are missing for both few-shot techniques due to the model's token limitations.

Table 7: Interquartile and median for a p-value of each group of journeys with similar size and different in-context prompting technique for gpt3.5-turbo-0125.

| | | 10 | 11 | 12 | 13 | 14 | 15 | 16 | 17 | 18 | 19 | 20 | 21 | 22 | 27 |
|---|---|---|---|---|---|---|---|---|---|---|---|---|---|---|---|
| Zero-shot | IQR | 0.2035 | 0.37375 | 0.29225 | 0.34625 | 0.42875 | 0.41675 | 0.43075 | 0.4045 | 0.3935 | 0.43825 | 0.395 | 0.43575 | 0.406 | 0.08025 |
| | Median | 0.4345 | 0.33000 | 0.41600 | 0.29500 | 0.19300 | 0.07050 | 0.33700 | 0.0935 | 0.3525 | 0.29550 | 0.336 | 0.16750 | 0.350 | 0.44850 |
| Zero-shot + CoT | IQR | 0.17575 | 0.4305 | 0.1330 | 0.2475 | 0.4205 | 0.42225 | 0.45925 | 0.3880 | 0.4025 | 0.4035 | 0.133 | 0.3735 | 0.3330 | 0.07825 |
| | Median | 0.43450 | 0.3940 | 0.4205 | 0.3985 | 0.4235 | 0.40250 | 0.39050 | 0.0515 | 0.2175 | 0.1420 | 0.449 | 0.0310 | 0.1615 | 0.44850 |
| Few-shot | IQR | 0.34425 | 0.18875 | 0.4320 | 0.4250 | 0.33575 | 0.3040 | 0.34175 | 0.35825 | 0.3635 | 0.34475 | 0.0670 | 0.0685 | None | None |
| | Median | 0.16650 | 0.09450 | 0.0615 | 0.1825 | 0.16750 | 0.0805 | 0.15800 | 0.04900 | 0.1085 | 0.23100 | 0.4505 | 0.4470 | None | None |
| Few-shot + CoT | IQR | 0.23275 | 0.09875 | 0.21775 | 0.32125 | 0.275 | 0.3655 | 0.3065 | 0.39925 | 0.2915 | 0.3435 | 0.0705 | 0.0825 | None | None |
| | Median | 0.08200 | 0.06300 | 0.07050 | 0.19900 | 0.091 | 0.1625 | 0.2395 | 0.16850 | 0.1955 | 0.2280 | 0.4390 | 0.3740 | None | None |

*D. Improving TSP Solution Quality with Self-Ensemble in Fine-Tuned GPT-3.5 Models*

In Figure 6, the graphs display gap statistics (i.e. Median and IQR) for TSP solutions from a fine-tuned GPT-3.5 model, showcasing the quality of the valid routes returned out of 30 instances evaluated at each size. Despite being fine-tuned on size-10 instances only, the model exhibits good generalisation to larger sizes. However, an increasing trend is noticeable as instance sizes increase beyond ten, indicated by rising gap median and interquartile range (IQR) values. Notably, implementing self-ensemble techniques shows a marked improvement in solution quality. This is evidenced by reduced median gap and IQR across all instance sizes. Moreover, the enhancement in trip quality is proportional to the ensemble size within the explored range, highlighting the benefit of self-ensemble in optimising the performance of the fine-tuned GPT-3.5 model for combinatorial tasks like the TSP.

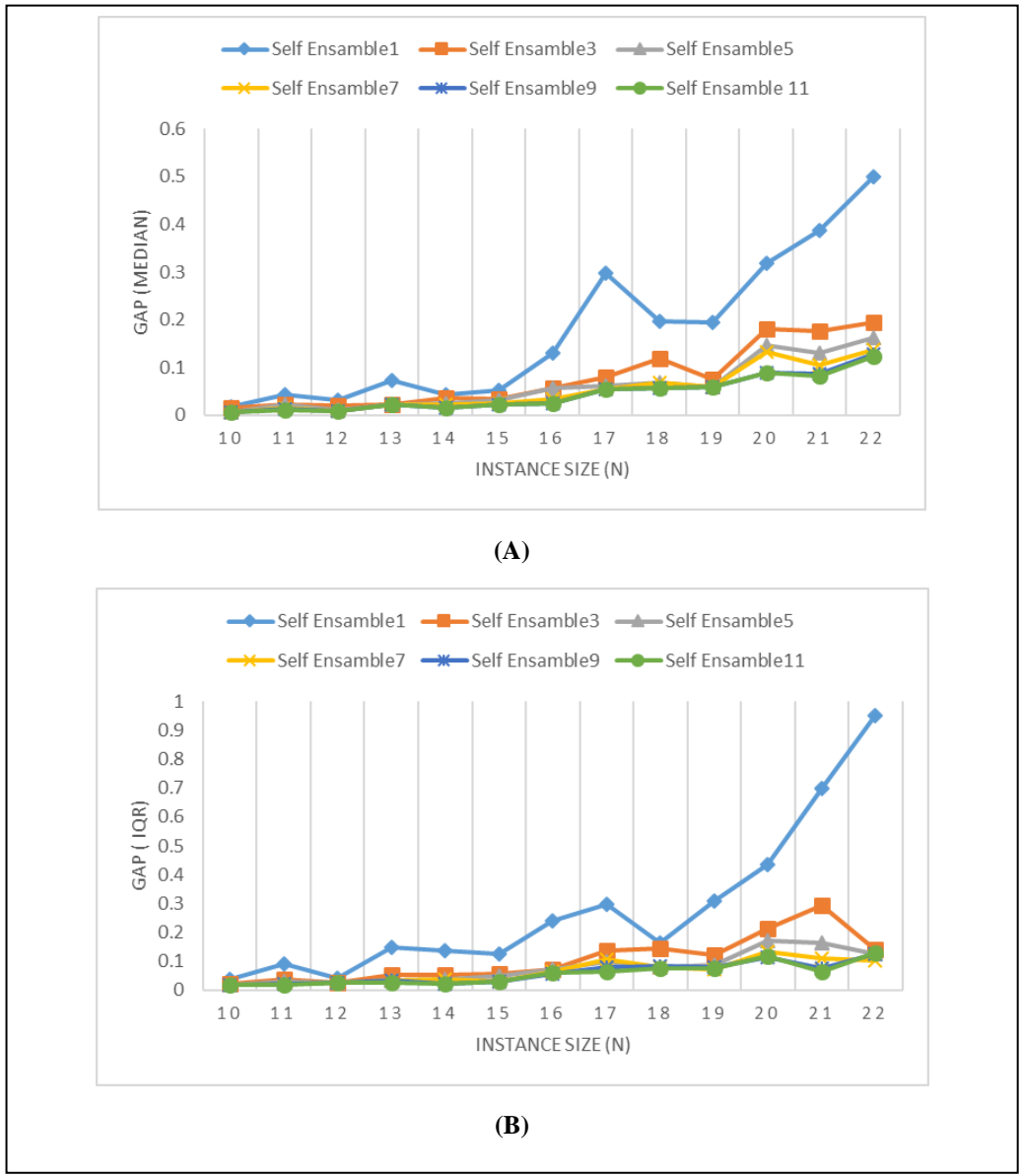

Figure 6: Median and interquartile for the gap of each group of journeys with similar size and different self-ensemble values for fine-tuned model

In Table 8, the analysis provides a detailed insight into the capabilities of the fine-tuned GPT-3.5 model in solving the Travelling Salesman Problem (TSP). As the size of the problem increases, both the median of the gap and the Interquartile Range (IQR) statistics tend to increase, yet the randomness scores suggest that the solutions are not by chance. This is particularly true for larger problems, where the big number of possible solutions makes the probability of generating a good one by chance very low. Even when there is a large gap statistic, most solutions are able to pass the randomness test, indicating that the model is effectively solving problems rather than randomly selecting the order of nodes.

In both presented cases in Figure 7, for 17 and 18-point TSP instances, the model's solutions exhibit a good degree of cost when measured against the optimal solutions. The paths generated by the model are close to the optimal journeys, maintaining a consistent pattern without making oscillating moves between far near nodes. Notably, the data points near the origin and the upper part of the graph are clustered, indicating a tight grouping of starting and ending points. The model's routes respect these clusters, proceeding systematically from point to point without erratic jumps that would suggest inefficiency or errors. This consistency points to the model's effectiveness in constructing logical and well-structured paths, validating the robustness of its problem-solving abilities for the TSP.

Table 8: Interquartile and median for a p-value of each group of journeys with similar size.

| | | 10 | 11 | 12 | 13 | 14 | 15 | 16 | 17 | 18 | 19 | 20 | 21 | 22 | 27 |
|---|---|---|---|---|---|---|---|---|---|---|---|---|---|---|---|
| Self-Ensemble 1 | IQR | 0.0065 | 0.0200 | 0.00275 | 0.00675 | 0.00375 | 0.001 | 0.00275 | 0.0035 | 0.001 | 0.001 | 0.00775 | 0.05975 | 0.23525 | 0.4475 |
| | Median | 0.0025 | 0.0035 | 0.00050 | 0.00100 | 0 | 0 | 0 | 0 | 0 | 0 | 0 | 0 | 0.00050 | 0.0055 |
| Self-Ensemble 3 | IQR | 0.00375 | 0.003 | 0.001 | 0 | 0 | 0 | 0 | 0 | 0 | 0 | 0 | 0 | 0 | 0.00425 |
| | Median | 0.00150 | 0.001 | 0 | 0 | 0 | 0 | 0 | 0 | 0 | 0 | 0 | 0 | 0 | 0 |
| Self-Ensemble 5 | IQR | 0.0030 | 0.00275 | 0 | 0 | 0 | 0 | 0 | 0 | 0 | 0 | 0 | 0 | 0 | 0.00075 |
| | Median | 0.0015 | 0 | 0 | 0 | 0 | 0 | 0 | 0 | 0 | 0 | 0 | 0 | 0 | 0 |
| Self-Ensemble 7 | IQR | 0.003 | 0.001 | 0 | 0 | 0 | 0 | 0 | 0 | 0 | 0 | 0 | 0 | 0 | 0 |
| | Median | 0.001 | 0 | 0 | 0 | 0 | 0 | 0 | 0 | 0 | 0 | 0 | 0 | 0 | 0 |
| Self-Ensemble 9 | IQR | 0.003 | 0.001 | 0 | 0 | 0 | 0 | 0 | 0 | 0 | 0 | 0 | 0 | 0 | 0 |
| | Median | 0.001 | 0 | 0 | 0 | 0 | 0 | 0 | 0 | 0 | 0 | 0 | 0 | 0 | 0 |
| Self-Ensemble 11 | IQR | 0.00275 | 0.001 | 0 | 0 | 0 | 0 | 0 | 0 | 0 | 0 | 0 | 0 | 0 | 0 |
| | Median | 0.00100 | 0 | 0 | 0 | 0 | 0 | 0 | 0 | 0 | 0 | 0 | 0 | 0 | 0 |

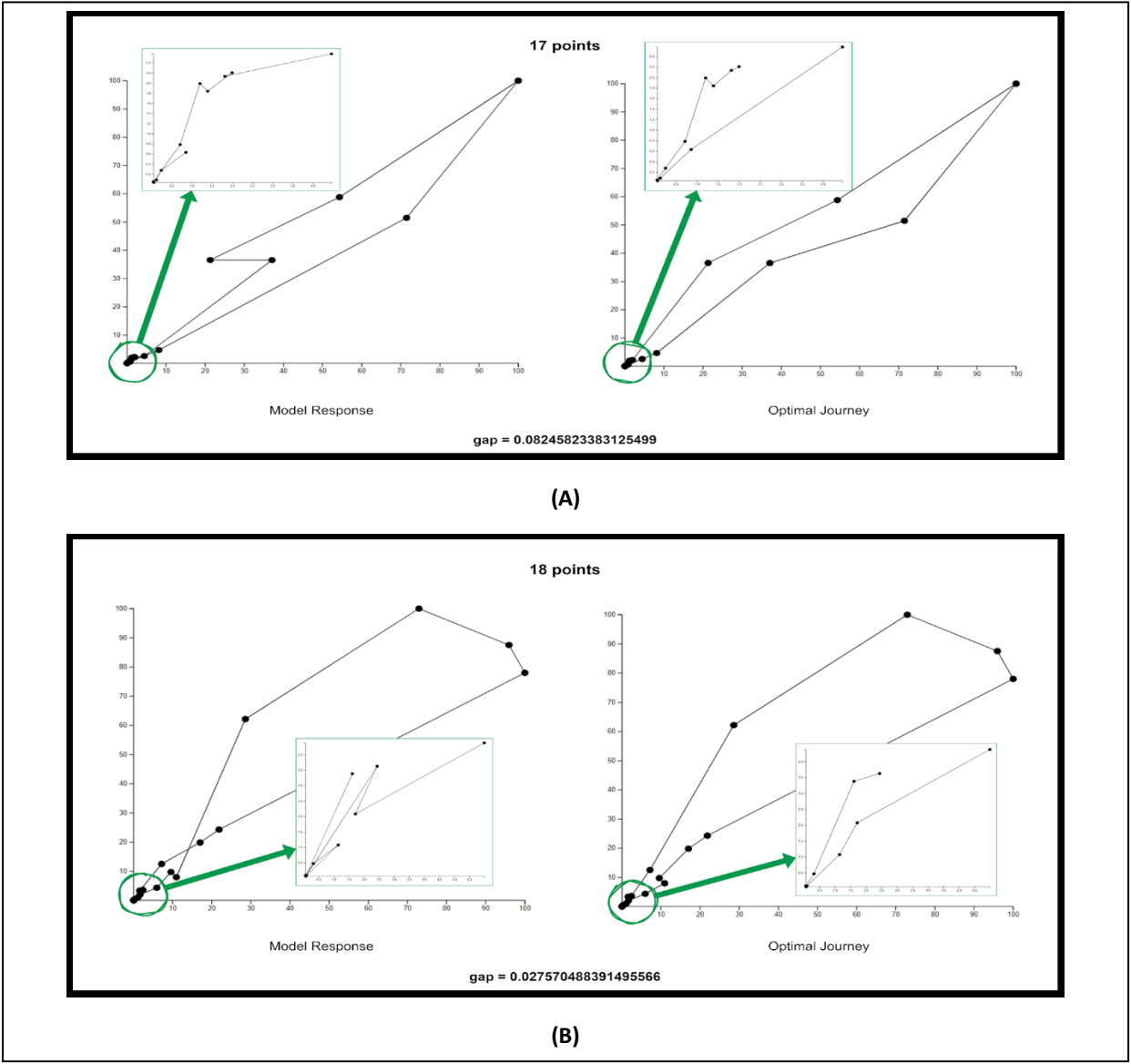

Figure 7: Model ordering for two instances with size 17 points, and 18 points

## 4. Conclusion and future work

This paper has explored the application of Large Language Models, particularly GPT-3.5 Turbo, to combinatorial problems, focusing on the Travelling Salesman Problem (TSP). Our findings indicate that LLMs are indeed capable of effectively solving TSP instances, utilizing prior knowledge in zero-shot scenarios and enhancing this with few-shot approaches, both with and without CoT processes. The model, when fine-tuned on fixed-size instances, showed impressive proficiency, especially with problems matching the training instances in size, and demonstrated considerable generalization capabilities to larger problems. The implementation of self-ensemble methods has further enhanced the model's performance, delivering improved solution quality without additional fine-tuning.

Future research should focus on refining the model's performance for larger instance sizes, where it currently falls short, potentially due to the increased complexity and token limits of the model. This should include advancements in prompt engineering to manage token count effectively and mitigate hallucinated responses, as well as exploring other open-source LLMs that might offer better efficiency. Another promising avenue is integrating evolutionary algorithms as an external optimization tool or employing the LLM itself to evolve solutions from self-ensemble outputs. Moreover, making the model more accessible to non-experts, particularly in small business settings, could democratise access to powerful computational tools. As LLMs continue to advance, we plan to investigate how an ensemble of quantized and fine-tuned models using fixed-size instances could impact overall robustness and efficiency in complex tasks, compared to models fine-tuned on variable-sized instances. Also, we plane to extend the research in using LLM to plan drones' journeys to solve logistics problems.

# Appendix A

Table 9: Interquartile and median for the gap of each group of journeys with similar size and different in-context prompting technique for gpt3.5-turbo-0125

|  |  | 10 | 11 | 12 | 13 | 14 | 15 | 16 | 17 | 18 | 19 | 20 | 21 | 22 | 27 |
|---|---|---|---|---|---|---|---|---|---|---|---|---|---|---|---|
| Zero-shot | IQR | 1.931482 | 0.845311 | 1.947055 | 1.277235 | 1.532004 | 1.246707 | 1.504980 | 1.715524 | 1.644068 | 1.386357 | 2.014804 | 1.613960 | 1.717422 | None |
| Zero-shot | Median | 0.443268 | 0.344298 | 0.538973 | 0.452925 | 0.492409 | 0.690858 | 0.845443 | 0.788674 | 0.994915 | 0.894570 | 1.323189 | 1.004896 | 1.276987 | None |
| Zero-shot + CoT | IQR | 2.008948 | 1.298093 | 2.745270 | 1.963786 | 1.215264 | 1.815348 | 1.596429 | 1.434338 | 1.777288 | 0.632359 | 1.782427 | 1.131237 | 1.187963 | None |
| Zero-shot + CoT | Median | 0.795407 | 0.193839 | 0.570557 | 0.601936 | 0.537696 | 0.481341 | 0.605338 | 0.778303 | 0.881434 | 1.018810 | 1.590040 | 0.846262 | 1.387670 | None |
| Few-shot | IQR | 1.565437 | 1.381503 | 2.080248 | 1.806493 | 1.071820 | 1.936998 | 1.582727 | 1.513339 | 1.449502 | 1.391533 | 1.750943 | 1.91286 | None | None |
| Few-shot | Median | 0.181104 | 0.066111 | 0.223686 | 0.512531 | 0.295425 | 0.369151 | 0.616459 | 0.468759 | 0.907383 | 1.015934 | 1.590040 | 1.57301 | None | None |
| Few-shot + CoT | IQR | 1.341473 | 0.845234 | 2.047197 | 1.348237 | 1.123594 | 1.755933 | 1.520379 | 1.070367 | 1.758138 | 1.173349 | 2.293085 | 1.451817 | None | None |
| Few-shot + CoT | Median | 0.123957 | 0.152856 | 0.577578 | 0.352176 | 0.490923 | 0.350027 | 0.837694 | 1.011812 | 1.093788 | 1.124133 | 0.913524 | 1.097716 | None | None |

# Appendix B

Table 10: Interquartile and median for the gap between response distance and optimal distance of each group of journeys with similar size

|  |  | 10 | 11 | 12 | 13 | 14 | 15 | 16 | 17 | 18 | 19 | 20 | 21 | 22 |
|---|---|---|---|---|---|---|---|---|---|---|---|---|---|---|
| Self-Ensemble 1 | IQR | 0.037895 | 0.091205 | 0.039042 | 0.147063 | 0.135863 | 0.126818 | 0.240842 | 0.297736 | 0.164906 | 0.309666 | 0.433997 | 0.697990 | 0.949996 |
| Self-Ensemble 1 | Median | 0.017989 | 0.043887 | 0.032278 | 0.072521 | 0.042544 | 0.051959 | 0.131262 | 0.297743 | 0.196158 | 0.195370 | 0.317500 | 0.386862 | 0.500586 |
| Self-Ensemble | IQR | 0.022755 | 0.038108 | 0.026783 | 0.053611 | 0.053472 | 0.055669 | 0.072431 | 0.135111 | 0.143487 | 0.123135 | 0.212135 | 0.292898 | 0.140230 |

| 3 | Median | 0.015602 | 0.022922 | 0.019149 | 0.023226 | 0.035572 | 0.032964 | 0.055946 | 0.080118 | 0.119427 | 0.075810 | 0.180414 | 0.175817 | 0.195264 |
|---|---|---|---|---|---|---|---|---|---|---|---|---|---|---|
| Self-Ensemble 5 | IQR | 0.017446 | 0.029675 | 0.023708 | 0.034448 | 0.038099 | 0.048635 | 0.072427 | 0.100089 | 0.077681 | 0.085771 | 0.170223 | 0.163529 | 0.123547 |
| | Median | 0.006906 | 0.021848 | 0.011523 | 0.023226 | 0.024301 | 0.031464 | 0.055946 | 0.060958 | 0.068568 | 0.058071 | 0.146117 | 0.130280 | 0.162183 |
| Self-Ensemble 7 | IQR | 0.017997 | 0.020364 | 0.024886 | 0.031765 | 0.036643 | 0.030605 | 0.062563 | 0.107118 | 0.080628 | 0.073032 | 0.132805 | 0.108371 | 0.102164 |
| | Median | 0.006906 | 0.015739 | 0.009586 | 0.022821 | 0.021594 | 0.023539 | 0.033454 | 0.055457 | 0.068426 | 0.058071 | 0.133226 | 0.105219 | 0.137977 |
| Self-Ensemble 9 | IQR | 0.01748 | 0.020277 | 0.025733 | 0.031765 | 0.023071 | 0.030255 | 0.057813 | 0.079649 | 0.084304 | 0.078737 | 0.115182 | 0.073711 | 0.126018 |
| | Median | 0.00595 | 0.012687 | 0.009586 | 0.022821 | 0.015145 | 0.021828 | 0.024950 | 0.055457 | 0.056083 | 0.058071 | 0.089404 | 0.086970 | 0.127638 |
| Self-Ensemble 11 | IQR | 0.018018 | 0.019479 | 0.025733 | 0.024722 | 0.023225 | 0.030255 | 0.058521 | 0.063430 | 0.076409 | 0.073673 | 0.117788 | 0.064988 | 0.13049 |
| | Median | 0.005797 | 0.011872 | 0.009586 | 0.022821 | 0.014822 | 0.021828 | 0.024950 | 0.054022 | 0.056083 | 0.058071 | 0.089404 | 0.081272 | 0.12357 |

## Appendix C

An example of an incomplete model response occurred with an instance of 27 points, where the model failed to provide the full sequence necessary for a valid journey due to reaching the token limit, resulting in an invalid and hallucinated trip.

Table 11: Testing in-context prompt with the fine-tunned response. It presents the possible hallucination due to model token size limit

---

**Prompts**

'In two-dimensional space, you will visit variable number of stations. You must visit each station once and return to the starting station at the end. Each station is represented with a 2-dimensional Cartesian point ( x, y ) where x is the coordinate on the X-axis and y is the coordinate on the Y-axis. The formula calculates the Euclidean distance between stations is " ( ( X1 - X2 ) ^ 2 + ( Y1 - Y1 ) ^ 2 ) ^ 0.5 ".

'Your task is to find the visiting order for the stations that minimizes the total distance you will travel to finish the journey.

"Let's work this out step-by-step to ensure we have the correct answer. First, calculate the Euclidean distance between all pairs of stations using the Euclidean distance formula and make a matrix of the calculated distances. Then, compare all the possible stations' orders to find the order that costs the minimum total travelling distance of the journey. At last, Sum the distances between the stations according to the order you find.

"The answer format should be as follows: distance matrix, Stations' order with minimum total traveling distance, and traveling distance.

station 0 ( 8884, 7468 ), station 1 ( 332113, 431807 ), station 2 ( 1290738, 2199389 ), station 3 ( 502198, 394047 ), station 4 ( 147901, 114385 ), station 5 ( 1238797, 973075 ), station 6 ( 476958, 450841 ), station 7 ( 457177, 477764 ), station 8 ( 1584324, 1719968 ), station 9 ( 369653, 596936 ), station 10 ( 3343988, 3361365 ), station 11 ( 1020190, 1099115 ), station 12 ( 4360676, 3454020 ), station 13 ( 1120200, 1224502 ), station 14 ( 1953928, 2373775 ), station 15 ( 7640, 9064 ), station 16 ( 30827, 30874 ), station 17 ( 1753716, 2226968 ), station 18 ( 4903787, 3285256 ), station 19 ( 3226558, 2970217 ), station 20 ( 7191825, 7354076 ), station 21 ( 5086, 6368 ), station 22 ( 22692, 29581 ), station 23 ( 14401038, 11535695 ), station 24 ( 10145, 12992 ), station 25 ( 1423780, 1430294 ), and station 26 ( 3601664, 2960096 ).

**Response**

{"distance_matrix":[["0.00","349061.98","1732368.82","495490.64","144653.72","1223630.34","478360.54","457584.54","1605040.03","382820.86","3365990.63","1024760.53","4295863.95","1080163.91","1972300.32","990.49","2353.73","1742590.36","4851164.75","3182303.10","8308198.20","705.80","1899.50","14402802.88","1021.47","1428914.01","3594251.14"],["349061.98","0.00","1024085.33","47425.42","222869.67","962073.65","143472.71","141079.90","1287129.52","187019.07","1293023.96","593100.08","1052392.65","654744.10","855950.19","348319.77","362243.90","1604225.30","1975211.69","946143.08","4033225.44","344358.01","485018.64","14178418.39","342844.23","1379565.56","1292504.72"],["1732368.82","1024085.33","0.00","1025913.39","1148531.28","1224746.1

9","858932.11","856835.51","538577.51","1014091.45","2067961.30","1138585.23","3151260.16","1309890.01","675299.56","1732347.50","1710480.73","529621.47","3667590.78","1967326.04","1006826.64","1732668.90","1711613.45","2171996.44","1732340.67","2219450.31"],["495490.64","47425.42","1025913.39","0.00","307220.63","824185.60","56000.97","44336.27","1128441.02","213937.10","1036160.08","681075.36","2895010.76","732736.07","1509489.12","488297.26","472467.86","1820333.35","4438273.58","2369847.84","9151434.10","480598.46","54455.76","14098393.43","484121.28","1090134.43","1567484.43"],["144653.72","222869.67","1148531.28","307220.63","0.00","1091252.26","365096.31","343799.04","1498751.85","487190.94","3222329.22","1257998.70","4273264.89","1152701.39","1950479.65","136379.76","155922.19","2021524.26","4937207.47","3090115.51","8061540.40","145314.58","217799.58","14402668.75","161361.85","1507641.68","2699897.78"],["1223630.34","962073.65","1224746.19","824185.60","1091252.26","0.00","786738.27","780335.82","757235.92","869605.77","2163050.04","252727.47","3159131.13","129729.31","741408.32","1223516.21","1210413.43","471489.89","3677286.15","1996958.25","6021805.94","1223613.07","1208256.96","12235873.42","1223597.27","185364.21","761243.57"],["478360.54","143472.71","858932.11","56000.97","365096.31","786738.27","0.00","30537.46","1129260.00","172679.03","1032191.67","582312.20","3939062.72","740980.16","1759142.25","482204.35","458990.29","1900033.08","4470925.50","2688770.48","7927434.14","484930.40","132163.25","13811520.71","489023.83","1072826.72","1217843.58"],["457584.54","141079.90","856835.51","44336.27","343799.04","780335.82","30537.46","0.00","1132628.26","165128.82","1023123.64","557444.34","3906121.91","722614.83","1774840.85","459253.86","430743.65","1950362.74","4484724.42","2672027.63","7927161.31","458829.37","126822.46","13757897.31","471585.11","1070670.18","1215135.68"],["1605040.03","1287129.52","538577.51","1128441.02","1498751.85","757235.92","1129260.00","1132628.26","0.00","1477598.89","2043925.38","673654.62","1864773.63","464921.18","377379.96","1605008.63","1615799.58","340048.96","3349322.83","1439484.02","2479161.95","1605015.75","1610261.94","14288820.36","1599403.51","778846.10","2055744.73"],["382820.86","187019.07","1014091.45","213937.10","487190.94","869605.77","172679.03","165128.82","1477598.89","0.00","1220711.18","571080.12","3150586.14","717004.81","1600732.07","386611.11","363431.98","1611257.33","4601218.05","2564566.00","7692808.00","383392.51","134894.16","13811356.06","380443.48","1127507.29","1535801.68"],["3365990.63","1293023.96","2067961.30","1036160.08","3222329.22","2163050.04","1032191.67","1023123.64","2043925.38","1220711.18","0.00","1360748.76","1278970.21","2322482.22","1702276.69","3365990.31","3353151.32","1555671.18","1600684.34","299946.39","1610506.79","3365990.85","3358157.33","2731216.32","3365961.46","2255243.74","650777.38"],["1024760.53","593100.08","1138585.23","681075.36","1257998.70","252727.47","582312.20","557444.34","673654.62","571080.12","1360748.76","0.00","3348483.08","392476.76","937644.37","1019905.36","999509.61","1332820.07","3913938.18","2133040.40","6174243.63","1024258.43","1100181.07","14159613.26","1014400.74","409174.34","1042380.26"],["4295863.95","1052392.65","3151260.16","2895010.76","4273264.89","3159131.13","3939062.72","3906121.91","1864773.63","3150586.14","1278970.21","3348483.08","0.00","3397851.66","1927402.73","4296032.66","4291834.18","3352137.55","717894.75","3179467.02","4750365.36","4295916.92","4296056.69","9941669.64","4295409.02","2855509.42","2652564.01"],["1080163.91","654744.10","1309890.01","732736.07","1152701.39","129729.31","740980.16","722614.83","464921.18","717004.81","2322482.22","392476.76","3397851.66","0.00","1224755.24","1078903.69","1059216.37","1140035.10","3426152.52","1683200.67","6605464.77","1080185.34","1152491.90","14107873.01","1072236.17","587271.67","1048364.07"],["1972300.32","855950.19","675299.56","1509489.12","1950479.65","741408.32","1759142.25","1774840.85","377379.96","1600732.07","1702276.69","937644.37","1927402.73","1224755.24","0.00","1972351.09","1959338.58","267888.32","3073450.23","1289153.44","5551004.86","1972300.32","1972300.32","10374879.79","1971979.86","1392863.82","1401219.86"],["990.49","348319.77","1732347.50","488297.26","136379.76","1223516.21","482204.35","459253.86","1605008.63","386611.11","3365990.31","1019905.36","4296032.66","1078903.69","1972351.09","0.00","58357.84","1753800.30","4861515.36","3192859.79","8308313.52","824.12","1755.38","14402803.51","805.92","1429093.01","3594544.83"],["2353.73","362243.90","1710480.73","472467.86","155922.19","1210413.43","458990.29","430743.65","1615799.58","363431.98","3353151.32","999509.61","4291834.18","1059216.37","1959338.58","58357.84","0.00","1733137.00","4830569.55","3162626.02","8279841.95","705.80","1959.50","14402760.52","1710.67","1430555.68","3586939.58"],["1742590.36","1604225.30","529621.47","1820333.35","2021524.26","471489.89","1900033.08","1950362.74","340048.96","1611257.33","1555671.18","1140035.10","3352137.55","1140035.10","267888.32","1753800.30","1733137.00","0.00","3205934.94","1374444.48","6467970.11","1732340.67","1710480.73","10374879.79","1732680.29","1326289.07","1268251.21"],["4851164.75","1975211.69","3667590.78","4438273.58","4937207.47","3677286.15","4470925.50","4484724.42","3349322.83","4601218.05","1600684.34","3426152.52","717894.75","3426152.52","3073450.23","4861515.36","4830569.55","3205934.94","0.00","1864868.55","1905855.75","4830569.55","4830609.32","1268264.03","4830460.67","3442938.64","4687288.18"],["3182303.10","946143.08","1967326.04","2369847.84","3090115.51","1996958.25","2688770.48","2672027.63","1439484.02","2564566.00","299946.39","2133040.40","3179467.02","1683200.67","1289153.44","3192859.79","3162626.02","1374444.48","1864868.55","0.00","4218175.65","3179467.02","3162626.02","10801796.75","3179467.02","1972300.32","1268264.03"],["8308198.20","4033225.44","1006826.64","9151434.10","8061540.40","6021805.94","7692808.00","7927161.31","2479161.95","7692808.00","1610506.79","5551004.86","4750365.36","5551004.86","8279841.95","8308313.52","8279841.95","6467970.11","1905855.75","4218175.65","0.00","8308303.46","8279841.95","6997548.66","8308233.22","6423511.41","7857737.86"],["705.80","344358.01","1711613.45","480598.46","145314.58","1223613.07","458829.37","458829.37","1605015.75","383392.51","3365990.85","1972300.32","4295916.92","1972300.32","705.80","824.12","705.80","1732340.67","4830569.55","3179467.02","8308303.46","0.00","150794.06","14402667.47","617.00","1428883.45","3594544.83"],["705.80","485018.64","1710613.68","54455.76","217799.58","1208256.96","126822.46","126822.46","1610261.94","134894.16","3358157.33","1972300.32","4296056.69","1972300.32","1959.50","1755.38","1959.50","1710480.73","4830609.32","3162626.02","8279841.95","150794.06","0.00","14305555.15","1683.38","1430555.68","3594544.83"],["14402802.88","14178418.39","2171996.44","1409834.43","14402668.75","12235873.42","13757897.31","13757897.31","14288820.36","13811356.06","2731216.32","10374879.79","9941669.64","10374879.79","10374879.79","14402803.51","14402760.52","10374879.79","1268264.03","10801796.75","6997548.66","14402667.47","14305555.15","0.00","14402723.42","14012175.08","14305273.84"],["1021.47","342844.23","1732340.67","484121.28","161361.85","1024760.53","471585.11","471585.11","1599403.51","380443.48","3365961.46","1072236.17","4295409.02","1072236.17","805.92","804.37","805.92","1732680.29","4830460.67","3179467.02","8308233.22","617.00","1683.38","14402723.42","0.00","1429084.40","3594544.83"],["1428914.01","1379565.56","2219450.31","1090134.43","1507641.68","185364.21","1070670.18","1070670.18","778846.10","1127507.29","2255243.74","587271.67","2855509.42","587271.67","1428883.45","1429093.01","1430555.68","1268264.03","3442938.64","1972300.32","6423511.41","1428883.45","1430555.68","14012175.08","1429084.40","0.00","1008701.54"],["3594251.14","1292504.72","2731216.32","1567484.43","2699897.78","761243.57","1217843.58","1215135.68","2055744.73","1535801.68","650777.38","1048364.07","2652564.01","1048364.07","3594544.83","3586939.58","3594544.83","1268264.03","4687288.18","1268264.03","7857737.86","3594544.83","3594544.83","14305273.84","3594544.83","1008701.54","0.00"]],"minimum_distance_order":[0,15,21,22,24

Appendix D

Visualization of some of the solved instances

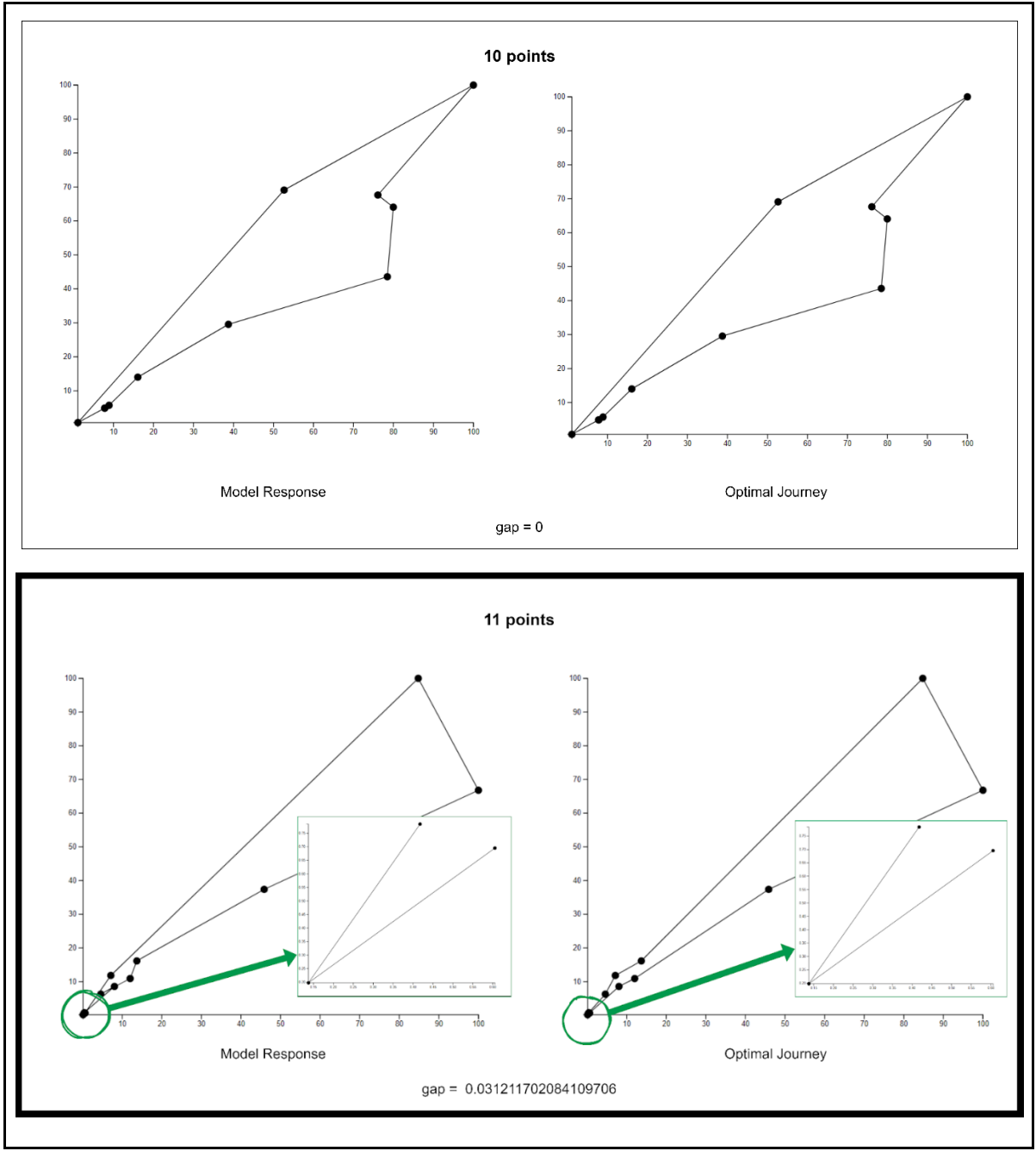

## 12 points

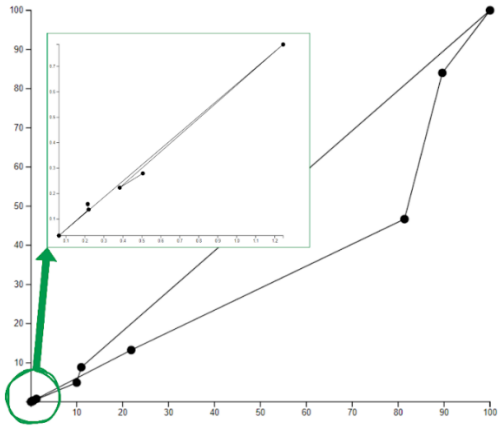
Model Response

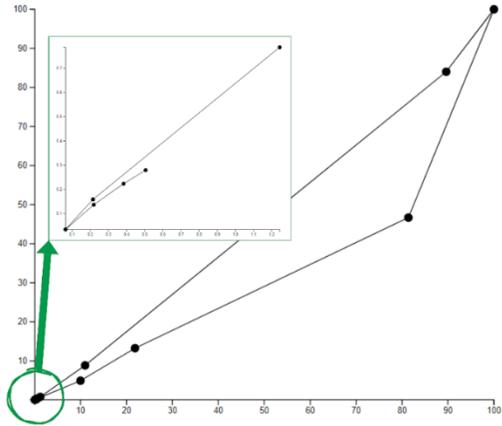
Optimal Journey

gap = 0.009464769648548633

## 13 points

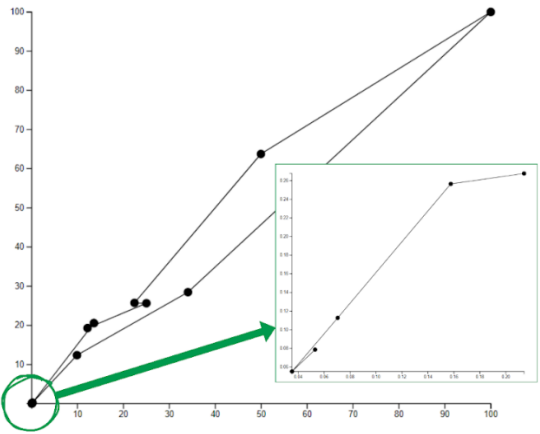
Model Response

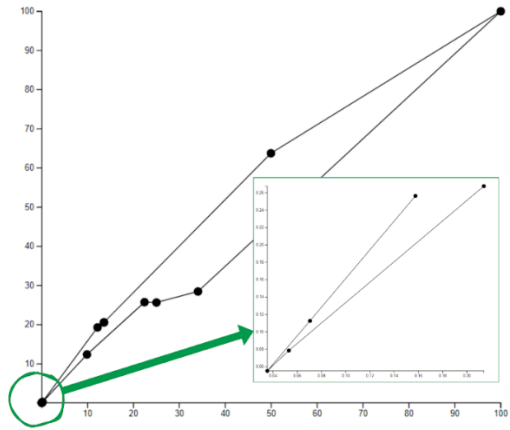
Optimal Journey

gap = 0.03065795158110272

## 14 points

Model Response | Optimal Journey

gap = 0.01621404464452178

## 15 points

Model Response | Optimal Journey

gap = 0.01779261350000784

## 16 points

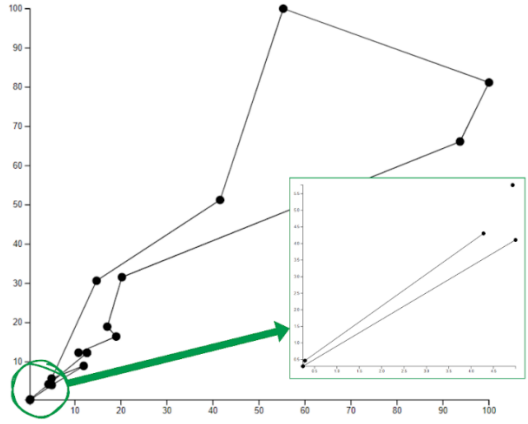
Model Response

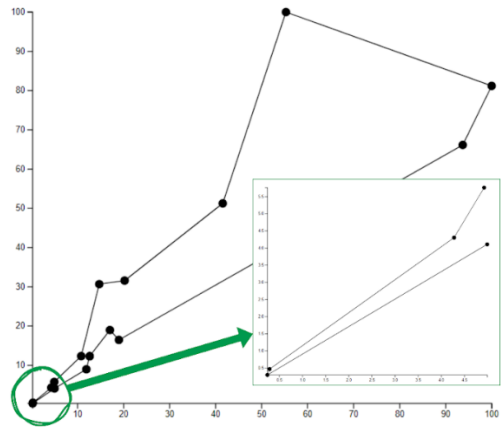
Optimal Journey

**gap = 0.07605220874817441**

## 19 points

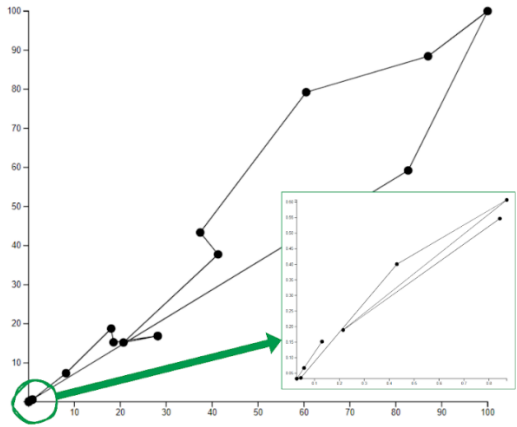
Model Response

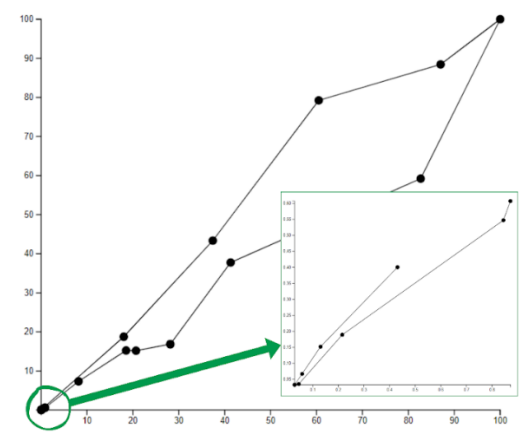
Optimal Journey

**gap = 0.08209591592779145**

## 20 points

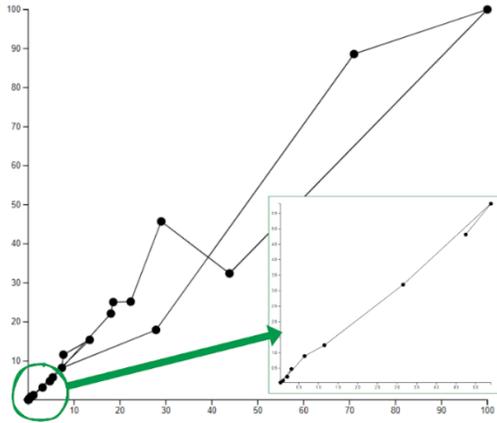
Model Response

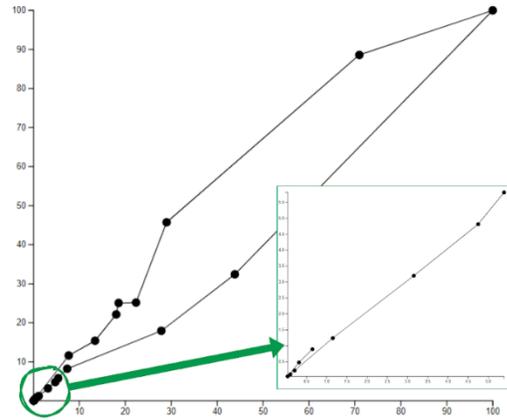
Optimal Journey

gap = 0.13865874105689768

## 21 points

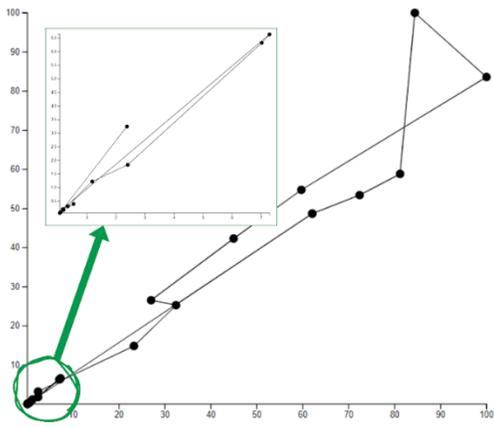
Model Response

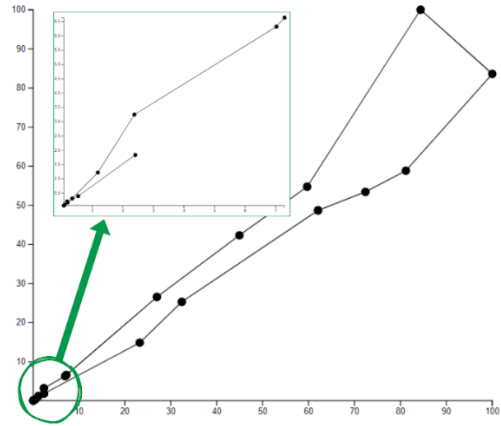
Optimal Journey

gap = 0.1145469183036806

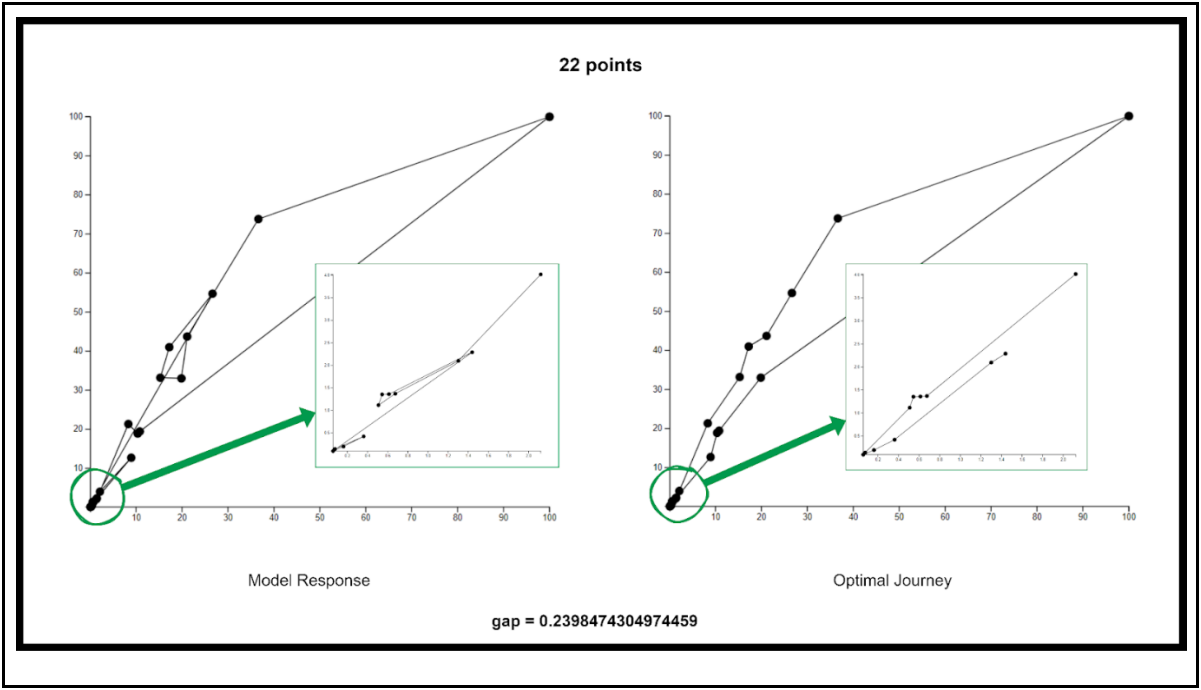

Figure 8: Optimal journey plan and model response's journey plan for journeys in different size groups